\documentclass[10pt,twocolumn,letterpaper]{article}

\usepackage[pagenumbers]{cvpr} % To force page numbers, e.g. for an arXiv version

\usepackage{flafter} 
\usepackage{float}
\usepackage{makecell}
\usepackage{multirow}
\usepackage{amsfonts,amssymb}
\usepackage{color, colortbl}
\usepackage{tabularx}
\usepackage[labelfont=bf]{caption}
\usepackage[accsupp]{axessibility}
\definecolor{cvprblue}{rgb}{0.21,0.49,0.74}
\usepackage[pagebackref,breaklinks,colorlinks,citecolor=cvprblue]{hyperref}
\def\paperID{2245} % *** Enter the Paper ID here
\def\confName{CVPR}
\def\confYear{2024}

\definecolor{green}{rgb}{0.0, 0.5, 0.0}
\definecolor{colorwheel}{rgb}{0.0, 0.5, 1.0}
\definecolor{ballblue}{rgb}{0.16, 0.32, 0.75}
\definecolor{LightCyan}{rgb}{0.88,1,1}
\DeclareMathOperator*{\argmax}{arg\,max}

\newcommand{\myparagraph}[1]{\vspace{2pt}\noindent{\bf{#1}}}

\title{Hyperbolic Learning with Synthetic Captions for Open-World Detection}

\author{
Fanjie Kong\textsuperscript{1}\thanks{Work done during an internship with AWS AI Labs.} , \hspace{3pt} Yanbei Chen\textsuperscript{2}\thanks{Corresponding author.} , \hspace{3pt} Jiarui Cai\textsuperscript{2}, \hspace{3pt} Davide Modolo\textsuperscript{2} \\
\textsuperscript{1}Duke University \hspace{2pt}
\textsuperscript{2}AWS AI Labs \\
{\tt\small fk43@duke.edu, $\;$ \{yanbec,cjiarui,dmodolo\}@amazon.com}
}

\begin{document}
\maketitle

\begin{abstract}

Open-world detection poses significant challenges, as it requires the detection of any object using either object class labels or free-form texts. Existing related works often use large-scale manual annotated caption datasets for training, which are extremely expensive to collect. Instead, we propose to transfer knowledge from vision-language models (VLMs) to enrich the open-vocabulary descriptions automatically. Specifically, we bootstrap dense synthetic captions using pre-trained VLMs to provide rich descriptions on different regions in images, and incorporate these captions to train a novel detector that generalizes to novel concepts. To mitigate the noise caused by hallucination in synthetic captions, we also propose a novel hyperbolic vision-language learning approach to impose a hierarchy between visual and caption embeddings. We call our detector ``HyperLearner''. We conduct extensive experiments on a wide variety of open-world detection benchmarks (COCO, LVIS, Object Detection in the Wild, RefCOCO) and our results show that our model consistently outperforms existing state-of-the-art methods, such as GLIP, GLIPv2 and Grounding DINO, when using the same backbone. 

\end{abstract}   

\section{Introduction}
\label{sec:intro}

\begin{figure}[!htb]
  \centering
    \includegraphics[width=0.46\textwidth]{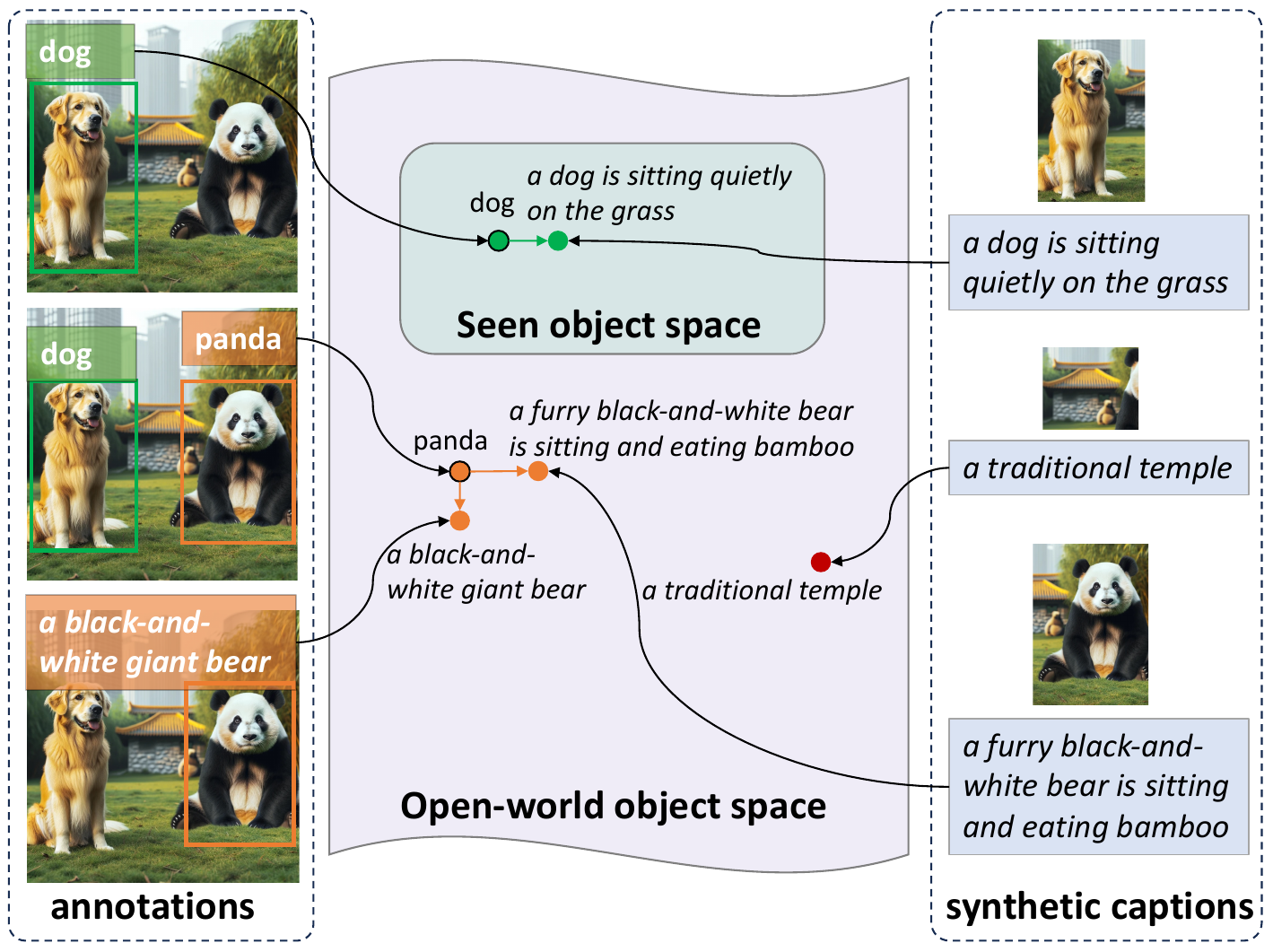}
   \vskip -0.5em
   \caption{
   We tackle the task of detecting seen and unseen objects using keywords (e.g., {\em panda}) or free-form texts (e.g., {\em a black-and-white giant bear}) in open world. We exploit synthetic captions from pre-trained caption models to bring rich open-world knowledge for training. As synthetic captions may be noisy, we propose to align visual features with text embeddings in a structural hierarchy to learn robustly and effectively from these captions.
   }
   \label{fig:teaser}
   \vskip -1em
\end{figure}

An intelligent perception model of our visual world requires the open-world generalizability of understanding any object. To enable such generalizability, recent advances in object detection integrate vision models with natural language models to enrich their open-world knowledge \cite{zareian2021open,zhou2022detecting,gu2021open,wu2023aligning,zhang2022glipv2,kuo2022findit}. 
Most of these works~\cite{zareian2021open,zhou2022detecting,gu2021open,wu2023aligning} consider the open-vocabulary setting in which a novel object is specified by its class label (e.g., {\em panda}). Other works \cite{zhang2022glipv2,kuo2022findit} cast detection as a generalized localization task and consider a more challenging open-set setting in which objects are described by free-form texts (e.g., {\em a black-and-white giant bear}). In this work, we tackle the task of detecting and localizing any object using both class labels and free-form texts (Figure \ref{fig:teaser}). We refer to this challenging task as {``open-world detection''}.

To improve generalization in open world, existing detection methods \cite{zareian2021open,li2022grounded,zhang2022glipv2} utilize language descriptions from large-scale image captioning datasets (e.g., MS-COCO captions \cite{lin2014microsoft}, conceptual captions \cite{sharma2018conceptual}) for training. These descriptions are however expensive to manually annotate \cite{lin2014microsoft} or design human-crafted data cleaning pipeline to process alt-text \cite{sharma2018conceptual}. To overcome this limitation, we propose to leverage recent advanced vision-language models to bootstrap machine-generated captions automatically. We collect these synthetic captions on different regions of an image to generate a dense understanding of all its concepts: from rich semantic information of known/seen objects (e.g., their attributes, states, actions and interactions) to diverse open-vocabulary descriptions of new concepts. Our intuition is that by aligning the detector features with these rich semantics, we can boost the model capability of understanding and recognizing any object, including novel ones. 

Nevertheless, one needs to be careful when using synthetic captions, as they are sometimes hallucinated or unrelated to the objects in the image \cite{rohrbach2018object}. For instance, given an image of a {\em panda} (Figure \ref{fig:teaser}), the model may generate the caption: ``{\em a furry black-and-white bear is sitting and eating bamboo}'', hallucinating the action ``{\em eating bamboo}'', which usually co-occurs with the animal {\em panda}. To mitigate the negative impact of aligning visual features with noisy captions, we propose to learn visual-textual alignment in a hierarchical structure, in which the synthetic caption entails the keyword of object, along with other semantics. We refer to this hierarchy structure as {\em `caption entails object'}. To impose such a structure in our representation space, we propose a novel hyperbolic vision-language learning objective which ensures that the caption embeddings entails the visual object embeddings in the representation space. Our formulation is inspired by the advances of structural representation learning with hyperbolic geometry \cite{dhingra2018embedding,khrulkov2020hyperbolic,ganea2018hyperbolic,ge2023hyperbolic,desai2023hyperbolic}. Rather than improving the unimodal representation learning on purely images \cite{ge2023hyperbolic} or texts \cite{dhingra2018embedding}, 
our hyperbolic learning objective is introduced for cross-modal learning on images and synthetic captions, and especially addresses the hallucination problem of synthetic captions to advance the open-world generalization in detection. 

\noindent To summarize, our contribution is three folds: 
\begin{itemize}[noitemsep,leftmargin=12pt,font=\bfseries]
\item[(1)] We propose to boost open-world detection using synthetic captions generated by powerful pre-trained VLMs. These synthetic captions provide rich language descriptions of both seen and unseen objects, enabling the detection of both new class labels and free-form texts. 
\item[(2)] We formulate a novel hyperbolic vision-language learning objective to help learning robust representation using noisy synthetic captions. Our hyperbolic loss formulation imposes a meaningful structural hierarchy in the representation space to specifically tackle the hallucination problem in these captions. 
\item[(3)] We show that our approach achieves the state-of-the-art performance on a variety of detection and localization datasets in the open-world setting, including COCO, LVIS, Object Detection in the Wild (ODiW), and RefCOCO. Our approach outperforms existing open-world detectors such as GLIP, GLIPv2, and Grounding DINO when using the same backbone and datasets for training. 
\end{itemize} 
    
\section{Related Work}
\label{sec:relatedwork}

\myparagraph{Open-vocabulary detection (OVD).} 
{Standard detection models like Faster R-CNN~\citep{fastrcnn} and DETR~\citep{carion2020end} succeed in detecting objects from a fixed vocabulary, but are not capable of localizing novel concepts.
To overcome this limitation, OVD methods take advantage of extensive image-text datasets through vision-language models (VLMs). 
OVD approaches follow two trends: learning novel object concepts from image-level supervision~\citep{zhou2022detecting}, and transferring knowledge from VLMs, which includes distilling visual-semantic cues~\cite{gu2021open,wu2023aligning}, aligning image-text representations~\cite{zang2022open,zhong2022regionclip,wu2023cora}, and exploiting pseudo-labels from captioning models~\cite{cho2023open,li2023blip}. 
In line with the more contemporary trend, our approach seeks to assimilate novel semantic concepts by probing VLMs. 
Specifically, we focus on distilling knowledge from large amount of (noisy) captions generated by these VLMs. 
Moreover, we also delve into the integration of free-form texts in open-world localization, lifting the model's proficiency in understanding complex referring expressions beyond plain object classes. }

\myparagraph{Vision-language models (VLMs) in detection.} 
VLMs have demonstrated remarkable abilities in tasks like image classification and image-text retrieval \cite{radford2021learning,desai2021virtex,jia2021scaling,yao2021filip,yang2022unified,zhai2022lit}. Recently, VLMs have also pioneered the state-of-the-art in detection, as represented by GLIP~\cite{li2022grounded,zhang2022glipv2}, RegionCLIP~\cite{zhong2022regionclip} and Grounding DINO~\cite{liu2023grounding}. These methods use large-scale image-text datasets to train the detector, thus enhancing the knowledge of both seen and unseen concepts. Concurrently, other studies~\cite{gu2021open, wu2023aligning,zhou2022detecting,chen2023scaledet} have focused on exploiting the visual-semantic knowledge from VLMs, e.g., Detic~\cite{zhou2022detecting} and ScaleDet~\cite{chen2023scaledet} use the text encoder from CLIP to encode the class labels of any seen and unseen objects. Our work contributes to this evolving direction with a novel hyperbolic vision-language learning approach to learn with synthetic captions from VLMs, which avoids the need of expensive manual annotated image-text datasets.

\myparagraph{Learning with captions.} Image captions convey rich information about objects in an image, such as object attributes, actions, relationships and scenes. Recent works have highlighted the advantages of harnessing caption data to enhance visual representation learning \cite{desai2021virtex,yu2022coca,tschannen2023image,yang2023alip,doveh2023dense}. The majority of these studies exploit {\em manual annotated captions}, training visual encoders alongside auxiliary caption generators to learn more generalized visual features \cite{desai2021virtex,yu2022coca,tschannen2023image}. 
A few studies investigate the use of {\em machine generated captions} to augment supervision for visual learning \cite{yang2023alip,doveh2023dense}. 
Our approach takes an innovative step further by employing synthetic captions to improve open-world generalization in detection. Importantly, we tackle the hallucination issue that often presents in synthetic captions by introducing a novel hyperbolic loss formulation.

\myparagraph{Hyperbolic learning.} 
Representation learning based on hyperbolic geometry has shown considerable success in various domains such as text embeddings~\cite{dhingra2018embedding}, text generation~\cite{dai2020apo} and knowledge graph modeling~\cite{chami2020low}, as the hyperbolic geometry could model the intrinsic hierarchy among data. Recently, a few works explore hyperbolic learning in vision tasks to model the hierarchical relation between objects and scenes \cite{ge2023hyperbolic}, or images and texts \cite{desai2023hyperbolic}. Our work is built on the foundation of hyperbolic learning to advance performance in detection. We specially delve into modelling the hierarchy between visual objects and synthetic captions, 
and formulate a novel hyperbolic learning objective to learn upon such hierarchy.

\begin{figure*}[!t]
    \centering
    \includegraphics[width=0.94\textwidth]{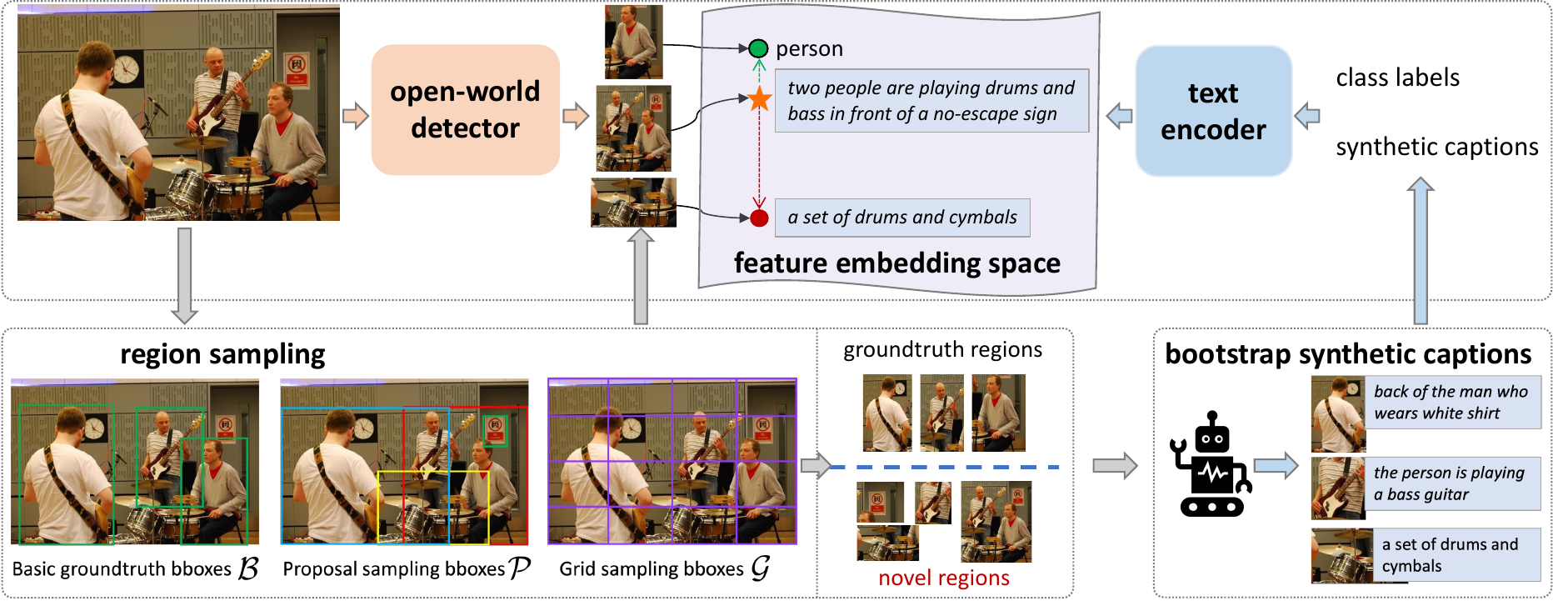}
    \vskip -0.8em
    \caption{{\bf Approach overview.} 
    Given an image, our open-world detector extracts visual feature embeddings on region crops, and aligns visual feature embeddings with text embeddings extracted from a pre-trained text encoder (\S \ref{sec:prelim}). To obtain synthetic captions for both seen and novel objects, we propose region sampling to augment region crops, and adopt a pre-trained image captioner to bootstrap synthetic captions on these region crops (\S \ref{sec:syn_caps}). To learn from synthetic captions effectively, we propose to align visual feature embeddings and caption embeddings in a structural hierarchy through hyperbolic vision-language learning in the hyperbolic space (see Figure \ref{fig:loss}, \S \ref{sec:hyp_learn}). 
    }
    \label{fig:model}
    \vskip -1.5em
\end{figure*}

\section{Proposed Approach}
\label{sec:method}

We first present the preliminaries of our model in \S \ref{sec:prelim}. We then introduce our caption bootstrapping strategy which provides rich open-vocabulary descriptions for training (\S \ref{sec:syn_caps}). Finally, we present our hyperbolic vision-language learning approach which exploits rich semantics from synthetic captions to boost open-world generalization (\S \ref{sec:hyp_learn}).

\subsection{Preliminaries on model architecture}\label{sec:prelim}

\myparagraph{Training an open-world detector.} Object detectors are trained to predict a list of bounding box location $B_i \in \mathbf{R}^4$ and class label $L_i \in \mathbb{R}^n$. They often consist of an visual encoder, a bounding box regressor and a visual classifier. During training, a bounding box regression loss $\mathcal{L}_{bbox}$ and a classification loss $\mathcal{L}_{cls}$ are jointly optimized to predict the object bounding box location and class label, i.e.,
\begin{equation}
\begin{aligned}
\mathcal{L}_{Det} = \mathcal{L}_{bbox} + \mathcal{L}_{cls}.
\end{aligned}
\label{eq:basic}
\end{equation}
Most detectors can be categorized into one-stage~\cite{tian2019fcos,zhou2019objects} or two-stage~\cite{ren2015faster,he2017mask,zhou2021probabilistic}. Detectors in the latter set employ an additional loss to train a region proposal network ($\mathcal{L}_{rpn}$) for predicting the objectness score of each region proposal. 
For the classification loss $\mathcal{L}_{cls}$, classic detectors often adopt a cross-entropy loss~\cite{ren2015faster}, which works well for classifying known classes, but it cannot model novel classes. To overcome this limitation, open-world detectors \cite{li2022grounded,zhou2022detecting} leverage text embeddings to represent the class labels. 
In details, the classification loss $\mathcal{L}_{cls}$ helps aligning the visual feature embedding $v_i$ with its class label embedding $l_i$, i.e., 
\begin{equation} 
\mathcal{L}_{cls} = - \log \frac{\exp(\text{sim}(v_i, l_i)/\tau)}{\sum_{j=1}^{n} \exp(\text{sim}(v_i, l_j)/\tau)},
\label{eq:cls}
\end{equation}
where $v_i$ is visual feature embedding from the detector, $\text{sim}(\cdot, \cdot)$ measures cosine similarity between embeddings, $\tau$ is the temperature. The class label embedding $l_j$ is extracted with a pre-trained text encoder from CLIP \cite{radford2021learning}. At test time, the predicted label is $L_i = \argmax_j \{\text{sim}(v_i, l_j)\}_{j=1}^n$, where $n$ is the number of pre-defined class labels (seen during training). One can augment these $n$ classes with novel classes (by adding their class label embeddings) to enable the open-set recognition on these novel concepts. 

\myparagraph{Cross-modal attention.} 
The above formulation (Eq. \eqref{eq:cls}) enables the detection of novel objects using their class label embeddings. However, it does not capture the spatial awareness of language semantics on different visual objects, thus fails to localize objects with more precise information, e.g., localizing {\em ``the person playing a bass guitar''} in Figure \ref{fig:model}.
To achieve open-world detection with free-form texts, we propose to inject the language semantics into the visual feature embedding $v_i$ by a cross-modal attention module~\cite{li2020unicoder, li2022grounded}. Specifically, given any text embeddings $T = \{t_j\}_{1}^n$, we compute the cross-modal attention between $v_i$, $T$ as:
\begin{equation}
\begin{aligned}
   & v^{(q)}_i = (W^{(q)})^\top v_i ,\ T^{(k)} = T W^{(k)} ,\ T^{(v)} = T W^{(v)}, \\
   & A_i = T^{(k)}v^{(q)}_i/\sqrt{d}, \ v_i^l = \text{softmax}(A_i)\; T^{(v)} W^{(out)},
\end{aligned}
\label{eq:attention}
\end{equation}
where $W^{(q)}$, $W^{(k)}$, $W^{(v)}$ are the trainable query, key, and value matrices in multi-head attention~\cite{vaswani2017attention}, $d$ is the dimension of hidden embeddings $v_i^{(q)}$ and $T^{(k)}$, and $v_i^l$ is the visual feature embedding fused with language semantics. 
To further inject spatial awareness into the visual feature embedding, we add a positional encoding layer \cite{dosovitskiy2020image} to fuse the visual feature embedding $v_i$ with region proposal feature $p_i$:
\begin{equation}
\begin{aligned}
   v_i^s = \text{PositionalEncoding}(v_i, p_i).
\end{aligned}
\label{eq:spatial}
\end{equation} 
Finally, to attain a spatial-aware visual feature embedding fused with language semantics, we combine the language-aware and spatial-aware visual embeddings $v_i^l$ (Eq. \eqref{eq:attention}), $v_i^s$ (Eq. \eqref{eq:spatial}) with the region proposal network $\text{RPN}(\cdot)$: 
\begin{equation}
\begin{aligned}
 \Tilde{v}_{i} = \text{RPN}(v_i^l + v_i^s).
\end{aligned}
\label{eq:fuse}
\end{equation} 
Within this formulation, our detector can localize objects given any texts by replacing the original visual embedding $v_i$ with the new visual embedding $\Tilde{v}_{i}$, and compute the localization score based on similarity between $\Tilde{v}_{i}$ and the text embeddings $\{t_j\}_{1}^n$, i.e., $Loc_i = \argmax_j \{\text{sim}(\Tilde{v}_{i}, t_j)\}_{j=1}^n$.

\subsection{Bootstrapping synthetic captions}\label{sec:syn_caps}

To enable generalization on novel objects, existing methods often exploit manual annotated datasets such as image classification \cite{ridnik2021imagenet}, captioning \cite{sharma2018conceptual} and grounding dataset \cite{krishna2017visual} to provide knowledge of novel concepts. Rather than relying on expensive manual annotations, we propose to bootstrap synthetic captions from a pre-trained VLM. Similar to manual annotations, these captions also provide rich open-vocabulary descriptions of novel concepts. 
In this paper, we use BLIP2 \cite{li2023blip} to generate captions -- which is trained on a large-scale dataset of 129M images. It is worth noting that our model is agnostic to the VLM used for caption generation, and thus can benefit from more powerful VLMs.

\myparagraph{Region sampling for caption bootstrapping.} We collect three sets of region proposals. First, for seen/known classes we use the ground truth bounding boxes $\mathcal{B}$. Then, to attain regions on novel objects, we propose to use two region sampling strategies: {\em proposal sampling} and {\em grid sampling}. In {\em proposal sampling}, we sample region proposals with high objectness scores and apply non-maximum suppression to remove duplicates; this leads to the set of regions $\mathcal{P}$. In {\em grid sampling}, we split the image into $ k \times k$ grids similar to ViT \cite{dosovitskiy2020image} to get a set of region crops $\mathcal{G}$. Together, these produce a rich set of region crops: $\mathcal{B} \cup \mathcal{P} \cup \mathcal{G}$ that encapsulate both seen and unseen objects. Finally, we generate synthetic captions $\mathcal{C}$ densely over these region crops with the BLIP2 captioner: 
\begin{equation}
\mathcal{C} = \text{Captioner}_{\text{BLIP2}}(\mathcal{B} \cup \mathcal{P} \cup \mathcal{G}),
\label{eq:caption}
\end{equation}
where the captions $\mathcal{C}$ are then used for model training. 

\begin{figure}[!t]
    \centering
    \includegraphics[width=.47\textwidth]{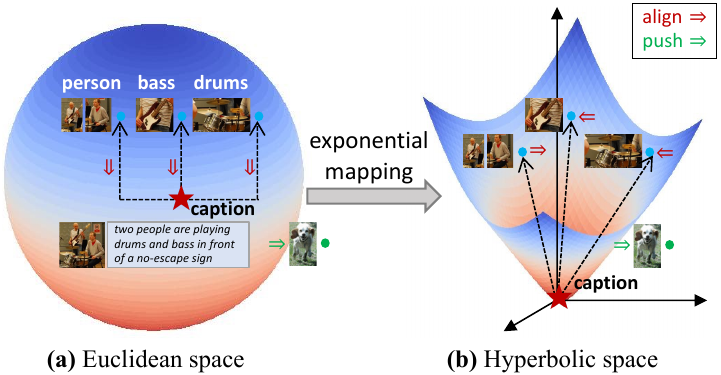}
    \vskip -0.8em
    \caption{
    {\bf Illustration of hyperbolic vision-language learning}. 
    The visual and caption embeddings are lifted from {\bf (a)} Euclidean space to {\bf (b)} Hyperbolic space by exponential mapping (Eq. \eqref{eq:hyp_exp}). To learn the partial order of {\em `caption entails object'}, we propose {hyperbolic contrastive loss}, {hyperbolic entailment loss} (Eq. \eqref{eq:hyp_contrastive}, Eq. \eqref{eq:entail}) to align visual and caption embeddings in hierarchy, thus ensuring the hallucination in caption is not aligned directly with visual embeddings to negatively impact the model learning. 
    }
    \vskip -1em
    \label{fig:loss}
\end{figure}

\subsection{Hyperbolic vision-language learning}\label{sec:hyp_learn}

\myparagraph{Vanilla contrastive learning.}
Our bootstrapping synthetic captions provide diverse language descriptions on both seen and unseen objects in different image regions, thus offering rich knowledge to boost open-world generalization. To learn from such knowledge, we can adopt the vanilla {\em contrastive loss} \cite{radford2021learning} for aligning the visual feature embedding $\Tilde{o}_{i}$ (Eq. \eqref{eq:fuse}) with the captions embedding $c_i$ attained in \eqref{eq:caption}: 
\begin{equation}
\mathcal{L}_{cap} = - \log \frac{\exp(\text{sim}(\Tilde{v}_i, c_i, )/\tau)}{\sum_{j=1}^{m} \exp(\text{sim}(\Tilde{v}_i, c_j)/\tau)},
\label{eq:contrastive}
\end{equation}
where $\mathcal{L}_{cap}$ is computed on a batch of $m$ caption embeddings, which enables learning about the novel concepts described in the synthetic captions. However, synthetic captions suffer from the hallucination problem~\cite{rohrbach2018object}, where concepts unrelated to the image are added to the generated caption; this simple contrastive loss is not equipped to handle them and its learning suffers from the generated noise. 
To learn more effectively from these captions, we propose to align visual and text embeddings in a hierarchy.

\myparagraph{Hyperbolic contrastive learning.} 
The synthetic caption on a region often entails the object descriptions along with other content.  
Thus, we argue that the object and caption should follow a {\em `caption entails object}' hierarchy in a tree-like structure (Figure \ref{fig:loss}). Inspired by recent advances in hyperbolic learning to model the partial order of {\em `scene entails object'} \cite{ge2023hyperbolic} or {\em `text entails image}' \cite{desai2023hyperbolic}, we propose to model our {\em `caption entails object}' hierarchy in a hyperbolic space. 

Specifically, we introduce our hyperbolic contrastive learning based on the Lorentzian distance in hyperbolic geometry \cite{law2019lorentzian}, which lifts the embeddings from Euclidean space to the Lorentz hyperboloid to represent the structural hierarchy. Formally, we first project the visual feature embedding $\Tilde{v}_i$ and caption embedding $c_i$ to the hyperboloid with exponential mapping (Eq. \eqref{eq:hyp_exp}), and then compute their Lorentzian distance on the hyperboloid (Eq. \eqref{eq:hyp_dist}) as follows:
\begin{equation}
\begin{aligned}
& v^{\mathcal{H}}_i = \text{expm}_{\mathbf{0}}(\Tilde{v}_i), \  c^{\mathcal{H}}_i = \text{expm}_{\mathbf{0}}(c_i), \\
& \text{with} \ \text{expm}_{\mathbf{0}}(x) = \frac{\text{sinh}(\sqrt{C}\lVert x \rVert)}{\sqrt{C}\lVert x \rVert}, 
\end{aligned}
\label{eq:hyp_exp}
\end{equation}
where $\text{expm}_{\mathbf{0}}(\cdot)$ is an exponential mapping, and $C$ is a learnable curvature parameter. Given two mapped embeddings $v^{\mathcal{H}}_i, c^{\mathcal{H}}_i$, we follow existing Lorentz model \cite{desai2023hyperbolic,nickel2018learning,le2019inferring,law2019lorentzian}, and compute the Lorentzian distance on the hyperboloid as:
\begin{equation}
\begin{aligned}
& d_{\mathcal{H}}(\Tilde{v}_i, c_i) = \sqrt{1/C} \cdot \text{cosh}^{-1}(-c \langle \Tilde{v}_i, c_i\rangle_{\mathcal{H}}), \\
& \text{with} \ \langle \Tilde{v}_i, c_i \rangle_{\mathcal{H}} {=} \langle \Tilde{v}_i, c_i \rangle - \sqrt{1/C + \lVert \Tilde{v}_i \rVert}\sqrt{1/C + \lVert c_i \rVert},
\end{aligned}
\label{eq:hyp_dist}
\end{equation}
where $\langle \cdot, \cdot \rangle_{\mathcal{H}}$ is the Lorentz inner product induced by the Riemannian metric of the Lorentz model \cite{desai2023hyperbolic,nickel2018learning,le2019inferring,law2019lorentzian}, and $d_{\mathcal{H}}(\cdot, \cdot)$ is the distance between two embeddings on the hyperboloid. By mappings the embeddings to the hyperbolic space, we derive the {\em hyperbolic contrastive loss} by replacing cosine similarity measure $\text{sim}(\cdot, \cdot)$ in Eq. \eqref{eq:contrastive} with the hyperbolic distance measure $d_{\mathcal{H}}(\cdot, \cdot)$, i.e,
\begin{equation}
 \mathcal{L}^{\mathcal{H}}_{cap} = -\log \frac{\exp(-d_{\mathcal{H}}(v^{\mathcal{H}}_i, c^{\mathcal{H}}_i)/\tau)}{\sum_{j=1}^{B}\exp(-d_{\mathcal{H}}(v^{\mathcal{H}}_i, c^{\mathcal{H}}_j)/\tau)},
\label{eq:hyp_contrastive}
\end{equation}
where $\mathcal{L}^{\mathcal{H}}_{cap}$ is our loss function to align visual embedding and caption embedding in the hyperbolic space.

\myparagraph{Hyperbolic entailment.} 
To impose our proposed structural hierarchy of {\em `caption entails object'}, we formulate a hyperbolic entailment loss based on the property of an {\em entailment cone} -- which defines a cone in the hyperbolic space to enforce partial order between embeddings \cite{le2019inferring}. Specifically, we define an entailment cone for caption embedding $c^{\mathcal{H}}_i$ as:
\begin{equation}
\begin{aligned}
A(c^{\mathcal{H}}_i) = \sin^{-1}(\frac{2K}{\sqrt{C}\lVert c^{\mathcal{H}}_i \rVert})
\end{aligned}
\label{eq:cone}
\end{equation}
where $K$ is a constant 0.1 used to avoid numerical overflow; $A(\cdot)$ is half aperture of the cone in hyperbolic space. With $A(c^{\mathcal{H}}_i)$, we introduce the entailment constraints to impose the hierarchy of {\em `caption entails object'}, by ensuring (1) the visual object embedding $v^{\mathcal{H}}_i$ lies inside the cone, and (2) other visual embeddings $v^{\mathcal{H}}_j$ ($j {\neq} i$) lie outside the cone. 

Formally, we denote the exterior angle between $v^{\mathcal{H}}_i$ and $c^{\mathcal{H}}_i$ as $\angle (c^{\mathcal{H}}_i, v^{\mathcal{H}}_i)$\footnote{Note: the exterior angle $\angle(a,b)$ in hyperbolic space is computed as $\angle(a,b)=\cos^{-1}(\frac{\sqrt{1/c + \lVert b \rVert}+\sqrt{1/c + \lVert a \rVert}c \langle a^{\mathcal{H}}_i,b^{\mathcal{H}}_i\rangle_{\mathcal{H}}}{\lVert a^{\mathcal{H}}_i \rVert \sqrt{c(\langle a^{\mathcal{H}}_i,b^{\mathcal{H}}_i \rangle)^2 - 1}})$, $c$ is curvature.}, and introduce a 
loss term to ensure the entailment constraint (1):
\begin{equation}
\begin{aligned}
& E(c^{\mathcal{H}}_i,v^{\mathcal{H}}_i) = \max (0, \angle(c^{\mathcal{H}}_i, v^{\mathcal{H}}_i) - A(c^{\mathcal{H}}_i)), 
\end{aligned}
\label{eq:first}
\end{equation}
where $E(\cdot, \cdot)$ ensures $\angle(c^{\mathcal{H}}_i, v^{\mathcal{H}}_i) {\leqslant} A(c^{\mathcal{H}}_i)$, thus enforcing $v^{\mathcal{H}}_i$ to lie inside the cone of $c^{\mathcal{H}}_i$. To further ensure both entailment constraints (1) and (2), we introduce the following {\em hyperbolic entailment loss} with two loss terms:
\begin{equation}
\begin{aligned}
\mathcal{L}_{\text{entail}} = E(c^{\mathcal{H}}_i,v^{\mathcal{H}}_i) + \sum_{j\neq i} \max(0, \gamma - E(c^{\mathcal{H}}_i,v^{\mathcal{H}}_j)),
\end{aligned}
\label{eq:entail}
\end{equation}
where the first term is Eq. \eqref{eq:first}; the second term is a max margin loss to ensure $\angle(c^{\mathcal{H}}_i, v^{\mathcal{H}}_j) {\geqslant} A(c^{\mathcal{H}}_i) + \gamma$, thus enforcing $v^{\mathcal{H}}_j$ to lie outside the cone of $c^{\mathcal{H}}_i$ with a margin $\gamma$.

\myparagraph{Overall objective.} 
Finally, given our proposed hyperbolic loss functions (Eq. \eqref{eq:contrastive}, Eq. \eqref{eq:entail}), we formulate our learning objective as follows: 
\begin{equation}
\mathcal{L}_{hyper} = \mathcal{L}_{bbox} + \mathcal{L}_{cls} + \mathcal{L}^{\mathcal{H}}_{cap} + \mathcal{L}_{\text{entail}}, 
\label{eq:ours}
\end{equation}
where $\mathcal{L}^{\mathcal{H}}_{cap}, \mathcal{L}_{\text{entail}}$ are our {hyperbolic contrastive loss}, and {hyperbolic entailment loss} to align visual and caption embedding in a structural hierarchy. An alternative to learn from synthetic captions is the vanilla contrastive loss (Eq. \eqref{eq:contrastive}), which leads to the baseline objective below: 
\begin{equation}
\mathcal{L}_{baseline} = \mathcal{L}_{bbox} + \mathcal{L}_{cls} + \mathcal{L}_{cap}.
\label{eq:baseline}
\end{equation}
We compare our {\bf Hyper}bolic {\bf Learn}ing objective $\mathcal{L}_{hyper}$ (Eq. \eqref{eq:ours}) to alternative objectives $\mathcal{L}_{Det}$ (Eq. \eqref{eq:basic}), $\mathcal{L}_{baseline}$ (Eq. \eqref{eq:baseline}) 
in ablation study to verify our design rationales. 
We refer our approach as ``{\bf HyperLearner}'' in experiments. 

\begin{table*}[!t]
    \centering
    \resizebox{0.82\textwidth}{!}{%
    \begin{tabular}{l @{\hskip 1mm}| l|c|c|c|c|c|c}
        \toprule
        & \multirow{2}{*}{Method} &  \multirow{2}{*}{Backbone}  &   \multirow{2}{*}{ \#Params}&   \multirow{2}{*}{FLOPs} & \multirow{2}{*}{\makecell{Pre-training Data}} &  \multicolumn{2}{c}{\makecell{COCO2017 val} }  \\ 
         & &   &  &  &  & Zero-shot &  Fine-tuning  \\ 
         \midrule
         1&Faster-RCNN ~\cite{fastrcnn} & RN50-FPN & 42M & 180G & COCO & - & 40.2 \\
         2&Faster-RCNN ~\cite{fastrcnn} & RN101-FPN & 54M &  313G & COCO & - & 42.0 \\
         3&Deformable DETR(DC5)~\cite{zhu2020deformable}& RN50 & 41M & 187G  & COCO & - & 41.1 \\
         4&CenterNetv2~\cite{zhou2021probabilistic} & RN50 & 76M & 288G & COCO & - & 42.9 \\
        \midrule
        5&Dyhead-T~\cite{dai2021dynamic} & Swin-T & 232M & 361G & O365 & 43.6 & 53.3 \\
        6&GLIP-T(A)~\cite{li2022grounded} & Swin-T & 232M & 488G & O365 &  42.9 & 52.9 \\ 
        7&GLIP-T(B)~\cite{li2022grounded} & Swin-T & 232M & 488G & O365 &  44.9 & 53.8 \\ 
        8&GLIP-T(C)~\cite{li2022grounded} & Swin-T & 232M & 488G & O365, GoldG & 46.7 & 55.1 \\
        9&DINO-T~\cite{zhang2022dino} & Swin-T &  - & - & O365 & 46.2 & 56.9 \\ 
        10&Grounding-DINO-T\textsuperscript{1}~\cite{liu2023grounding} & Swin-T & 172M & 464G & O365 & 46.7 & 56.9  \\
        11&Grounding-DINO-T\textsuperscript{2}~\cite{liu2023grounding} & Swin-T & 172M & 464G & O365, GoldG & 48.1 & 57.1 \\
        12&Grounding-DINO-T\textsuperscript{3}~\cite{liu2023grounding} & Swin-T & 172M & 464G & O365, GoldG, Cap4M & 48.4 & 57.2 \\
        \midrule
        13&HyperLearner (Ours) & Swin-T & \textbf{90M} & \textbf{324G} & O365 & \bf 47.6 & \bf 56.8 \\
        14&HyperLearner (Ours) & Swin-T & \textbf{90M} & \textbf{324G} & O365, GoldG & \textbf{48.4} & \textbf{57.4} \\ 
        \bottomrule
    \end{tabular}
    }
    \vskip -0.8em
    \caption{{\bf Comparison on COCO benchmark.}
    Results are given on both zero-shot and fine-tuning settings. Metric: mAP.}
    \vskip -1em
\label{tab:COCO}
\end{table*}
\begin{table}[!t]
    \centering
    \resizebox{0.49\textwidth}{!}{%
    \begin{tabular}{l|c|c|c}
    \toprule
    \multirow{2}{*}{Method} & \multirow{2}{*}{\makecell{Pre-training Data}} &  \multicolumn{2}{c}{\makecell{LVIS minival} }  \\ 
    & & AP & APr $|$ APc $|$ APf \\ 
    \midrule
    MDETR~\cite{kamath2021mdetr} & GoldG, RefCOCO & 24.2 & 20.9 $|$ 24.9 $|$ 24.3 \\ 
    DETCLIP-T~\cite{yao2022detclip} & O365 & 28.8 &  26.0 $|$ 28.0 $|$ 30.0 \\
    GLIP-T (C)~\cite{li2022grounded}  & O365, GoldG & 24.9 & 17.7 $|$ 19.5 $|$ 31.0 \\ 
    GLIP-T~\cite{li2022grounded} & O365, GoldG, Cap4M & 26.0  & 20.8 $|$ 21.4 $|$ 31.0 \\ 
    Grounding-DINO-T~\cite{liu2023grounding}  & O365, GoldG & 25.6& 14.4 $|$ 19.6 $|$ {32.2} \\ 
    Grounding-DINO-T~\cite{liu2023grounding} & O365, GoldG, Cap4M & {27.4} & {20.8 $|$ 21.4 $|$ 31.0} \\ 
    \midrule
    HyperLearner (Ours) & O365 & 25.5 & 25.9 $|$ 27.5 $|$ 23.7 \\
    HyperLearner (Ours) & O365, GoldG & \textbf{31.3} & \textbf{30.7 $|$ 32.6 $|$ 30.3} \\
    \bottomrule
    \end{tabular}
    }
    \vskip -0.8em
    \caption{{\bf Comparison on LVIS benchmark.} Metric: mAP.}
    \vskip -1em
\label{tab:LVIS}
\end{table}

\section{Experiments}
\label{sec:experiments}

In the following, we detail our experimental settings (\S \ref{sec:setup}) and provide thorough evaluation of our approach on a wide set of benchmark datasets in comparison to the state-of-the-arts (\S \ref{sec:sota}). Finally, we provide insightful ablation study (\S \ref{sec:ablation}) and qualitative results (\S \ref{sec:qualitative}) to analyze our approach. 

\subsection{Experimental settings}
\label{sec:setup}

\myparagraph{Implementation details.} Our approach utilizes CenterNetV2~\cite{zhou2022detecting} as the first-stage region proposal network, succeeded by a cross-modal attention module as detailed in \S \ref{sec:prelim}. Our text encoder is the 12-layer text transformer from CLIP ViT/B16~\cite{radford2021learning}. We used BLIP2 6.7b~\cite{li2023blip} to generate synthetic captions on region crops offline (\S \ref{sec:syn_caps}). For model pre-training, we used two datasets: Object 365 (O365)~\cite{shao2019objects365} and GoldG~\cite{kamath2021mdetr}, to ensure a fair comparison with other methods on the same evaluation benchmark datasets. In all experiments, we use Swin-Tiny~\cite{liu2021swin} as the backbone, and undergo training for 100,000 iterations with a batch size of 64 and a learning rate 0.0001. The initial 10,000 iterations serve as a warm-up phase, which is essential for optimal convergence. 
We train our model on 8 A100 GPUs. 
More details are provided in the supplementary material. 

\myparagraph{Object detection benchmark datasets.} 
We evaluate the open-world detection performance of our method on the most popular object detection datasets: (1) COCO~\cite{lin2014microsoft}, which has 80 common object classes, and (2) LVIS~\cite{gupta2019lvis}, which has 1203 object classes. Furthermore, we also evaluate on Object Detection in the Wild (ODinW)~\cite{li2022grounded} -- an assembly of 13 datasets, each one representing a different real-world fine-grained domain.  
Following the common practices, for zero-shot setting we evaluate our pre-trained models directly, while for fine-tuning setting, we fine-tune our pre-trained models on the target dataset. We report mean Average Precision (mAP) at IoU threshold of 0.5.

\myparagraph{Object localization benchmark datasets.} 
We evaluate our method on referring expression comprehension
which aims to locate novel objects using natural language expressions. 
We employ the following benchmark datasets: 
RefCOCO~\cite{kazemzadeh2014referitgame}, RefCOCO+~\cite{yu2016modeling} and RefCOCOg~\cite{mao2016generation} (collectively referred as RefCOCO/+/g) 
to evaluate localizing the objects given descriptive free-form texts. 
All three datasets utilize the images from the original COCO dataset; however, they differ in their textual descriptions: 
RefCOCO+ emphasizes purely on appearance, excluding spatial reference found in RefCOCO; 
RefCOCOg provides more comprehensive descriptions in sentences rather than phases.
Our evaluation adheres to the established evaluation protocol,  employing top-1 accuracy to determine the successful localization of objects based on the description, in line with the metric used in previous works \cite{yu2018mattnet,kamath2021mdetr,liu2023grounding}.

\begin{table*}[]
    \centering
    \resizebox{0.9\textwidth}{!}{%
    \begin{tabular}{l @{\hskip 1mm}|l|c|c|ccc|ccc|cc}
  \toprule
    &\multirow{2}{*}{Method}  & \multirow{2}{*}{\makecell{Pre-training Data}}& \multirow{2}{*}{\makecell{Fine-tuning}}  &  \multicolumn{3}{c}{\makecell{RefCOCO} } & \multicolumn{3}{c}{\makecell{RefCOCO+} } & \multicolumn{2}{c}{\makecell{RefCOCOg} }  \\ 
    & &  &  &  val & testA & testB & val & testA & testB & val & test \\
    \midrule
    1&CNN-LSTM~\cite{mao2016generation} & RefC & - & - & 63.15 & 64.21  & - & 48.73 & 42.13 & 62.14 & - \\
    2&MAttNet~\cite{yu2018mattnet} & RefC &  \checkmark & 76.65 & 81.14 & 69.99 & 65.33 & 71.62 & 56.02 & 66.58 & 67.27 \\ 
    3&RefTR~\cite{li2021referring} & VG & \checkmark & 85.65 & 88.73 &  81.16 & 77.55 & 82.26 & 68.99&  79.25&  80.01 \\ 
    4&MDETR~\cite{kamath2021mdetr} & GoldG,RefC & \checkmark & 86.75 & 89.58 & 81.41 & 79.52 & 84.09 & 70.62 & 81.64 & 80.89 \\
    5&DQ-DETR~\cite{liu2023dq} & GoldG,RefC & \checkmark  & 88.63 & 91.04 & 83.51 & 81.66 & 86.15 & 73.21 & 82.76 & 83.44\\
    \midrule
    6&GLIP-T(B)~\cite{li2022grounded} & O365,GoldG & \texttimes & 49.96 & 54.69 & 43.06 &  49.01 & 53.44 & 43.42 &  65.58 &  66.08\\ 
    7&GLIP-T(C)~\cite{li2022grounded} & O365,GoldG,Cap4M & \texttimes & 50.42 & 54.30 & 43.83&  49.50 & 52.78 &  44.59 & 66.09 & 66.89 \\ 
    8&Grounding-DINO-T~\cite{liu2023grounding} & O365,GoldG & \texttimes & 50.41 & 57.24 & 43.21 & 51.40 &  57.59&  45.81 & 67.46 & 67.13 \\ 
    9&Grounding-DINO-T~\cite{liu2023grounding} & O365,GoldG,RefC & \texttimes & 73.98 & 74.88 & 59.29 & 66.81&  69.91&  56.09 & 71.06 & 72.07 \\ 
    10&Grounding-DINO-T~\cite{liu2023grounding} & O365,GoldG,RefC & \checkmark & 89.19 & 91.86 & \bf 85.99 & 81.09 & \bf 87.40 & \bf 74.71 & \bf 84.15 & \bf 84.94 \\ 
    \midrule
    11&HyperLearner (Ours) & O365,GoldG & \texttimes & \bf 50.66 & \bf 60.87 & \bf 44.66 & \bf 59.29 &\bf 62.29 & \bf 45.43 & \bf 67.02 & \bf 67.44\\ 
    12&HyperLearner (Ours) & O365,GoldG,RefC & \texttimes & \bf77.89 & \bf76.92 &\bf 72.99 &\bf 67.54 & \bf75.55 & \bf57.54 & \bf77.00 & \bf76.79 \\ 
    13&HyperLearner (Ours) &  O365,GoldG,RefC & \checkmark & \bf 90.74  &\bf 92.09 & 85.46 &\bf 82.35 & 84.70 & 72.64 & 82.53 & 82.39 \\ 
    \bottomrule
    \end{tabular}
    }
    \vskip -0.8em
    \caption{{\bf Comparison on RefCOCO/+/g benchmark.} Metric: Top-1 accuracy.}
    \vskip -1em
\label{tab:refCOCO}
\end{table*}
\begin{table}[t]
    \centering
    \resizebox{0.49\textwidth}{!}{%
    \begin{tabular}{l|c|c|c|c}
    \toprule
       \multirow{2}{*}{Method} &  \multirow{2}{*}{Backbone} & \multirow{2}{*}{\makecell{Pre-training Data}} &  \multicolumn{2}{c}{\makecell{Test $\text{AP}_{avg}$} }  \\ 
    & & & zero-shot & full-shot \\ 
    \midrule
    Detic-R~\cite{zhou2022detecting} & RN50 & LVIS, COCO, IN-21K & 29.4 & 64.4 \\
    Detic-B~\cite{zhou2022detecting} & Swin-B & LVIS, COCO, IN-21K & 38.7 & 70.1 \\ 
    \midrule
    GLIP-T(A)~\cite{li2022grounded} & Swin-T & O365 & 28.7 & 63.6 \\ 
    GLIP-T(B)~\cite{li2022grounded}& Swin-T & O365 & 33.2 & 62.7 \\ 
    GLIP-T(C)~\cite{li2022grounded}& Swin-T & O365, GoldG & 44.4 & 63.9 \\ 
    Grounding-DINO-T~\cite{liu2023grounding} & Swin-T& O365, GoldG,Cap4M  &  44.9 & -\\
    \midrule
    HyperLearner (Ours) & Swin-T & O365 & 37.9 & 66.7 \\ 
    HyperLearner (Ours) & Swin-T& O365, GoldG & \textbf{45.2} & \textbf{68.9} \\ 
    \bottomrule
    \end{tabular}
    }
    \vskip -0.8em
    \caption{{\bf Comparison on ODinW benchmark.} Metric: mAP.}
    \vskip -1.5em
\label{tab:ODinW}
\end{table}

\subsection{Comparison to SOTA methods}
\label{sec:sota}

\myparagraph{Comparison on COCO.} 
In Table~\ref{tab:COCO}, we compare our HyperLearner to several recent models:
(1) traditional two-stage object detectors, including Faster-RCNN~\cite{ren2015faster}, Deformable DETR~\cite{zhu2020deformable} and CenterNetV2~\cite{zhou2022detecting}; 
(2) closed-set detectors which are adaptable for evaluation on COCO through class mapping between O365 and COCO, including Dyhead~\cite{dai2021dynamic} and DINO~\cite{zhang2022dino}; 
(3) open-world detectors including GLIP~\cite{li2022grounded} and Grounding DINO~\cite{liu2023grounding}, which not only use image-text data for pre-training, but also utilize model architectures to fuse visual-language representations. 

Table~\ref{tab:COCO} details the comparison on model efficiency and performance on COCO. 
As shown, our model has much smaller parameter size and computation cost (measured in FLOPs), which significantly improve model efficiency for real-world deployment. 
Moreover, while being lightweight, our model achieves better performance in both zero-shot and fine-tuning settings, with an mAP of 48.4 and 57.4 as compared to 48.1 and 57.1 obtained by the best competitor Grounding DINO-T when using O365, GoldG for training. Notably, our model pre-trained on O365, GoldG (in row 14) obtains similar performance as Grounding DINO-T pre-trained on O365, GoldG and Cap4M, an additional large-scale manually-annotated caption dataset \cite{li2022grounded} (row 12). This clearly shows how powerful machine-generated captions can be when used properly in HyperLearner. 

\myparagraph{Comparison on LVIS.} 
Table~\ref{tab:LVIS} shows the results on LVIS minival validation set. 
Our model demonstrates a significant advantage over GLIP and Grounding DINO on detecting novel rare classes -- outperforming them by +13.0\% and +16.3\% in APr respectively, when pre-trained on O365 and GoldG.
Our improved gain is maintained at 9.9\% even when other methods benefit from the additional Cap4M dataset. 
The results suggest our proposed method's capability to detect novel and rare concepts, attributing to the effective incorporation of semantic knowledge via bootstrapping synthetic captions and the vision-language alignments brought by our novel hyperbolic loss formulation. 

\myparagraph{Comparison on ODinW.}
{The ODinW dataset exemplifies more fine-grained, challenging real-world application scenarios. 
In Table~\ref{tab:ODinW}, we report the average AP over its 13 subsets.
Our finding aligns with those from the COCO and LVIS datasets, indicating that our model performs well in  zero-shot testing scenario. 
It outperforms GLIP by +4.7\% with pre-training solely on O365 and +0.8\% when GoldG is added. 
While our model shows the commendable zero-shot transferability, it also excels in the full-shot setting, suggesting the pre-trained model possesses strong generalizablity.}

\myparagraph{Comparison on RefCOCO/+/g.} 
Table~\ref{tab:refCOCO} shows the results of localizing objects with referring expressions. 
When pre-trained on O365, GoldG or pre-trained on O365, GoldG, RefC, our model (in row 11, 12) surpasses Grounding DINO (in row 8, 9) and GLIP (in row 6). With further fine-tuning on RefC, our model (in row 13) performs competitively compared to Grounding DINO (in row 10). The competitive results in Table \ref{tab:refCOCO} can be attributed to learning with our bootstrapped captions, which often include descriptions of object attributes and their spatial relationships that also exist in the natural language expressions in RefCOCO/+/g.

\begin{table}[!t]
    \centering
    \resizebox{0.39\textwidth}{!}{%
    \begin{tabular}{l @{\hskip 1mm}|l|c|cccc|c}
        \toprule
        & \multirow{2}{*}{Method} &  \multirow{2}{*}{COCO} & \multicolumn{4}{c|}{LVIS} & \multirow{2}{*}{ODinW} \\ 
        & & & AP & APr & APc & APf & \\ \midrule
        \midrule
        1&$\mathcal{L}_{Det}$ & 42.3 & 13.3 & 8.3 & 11.9 & 15.4 & 26.3 \\
        2&$\mathcal{L}_{baseline}$ & 46.8 & 23.6 & 22.7 & 24.3 & 24.6 & 38.6 \\ 
        \rowcolor{LightCyan} 
        3&$\mathcal{L}_{hyper}$ & \textbf{48.4} & \textbf{31.3} & \textbf{30.7} & \textbf{32.6} & \textbf{30.3} & \textbf{45.2} \\ 
        \bottomrule
    \end{tabular}}
    \vskip -0.8em
    \caption{{\bf Ablation study on learning objectives.} Metric: mAP.}
    \vskip -1.5em
\label{tab:ab_loss}
\end{table}
\begin{figure}[!t]
    \centering
    \includegraphics[width=0.83\linewidth]{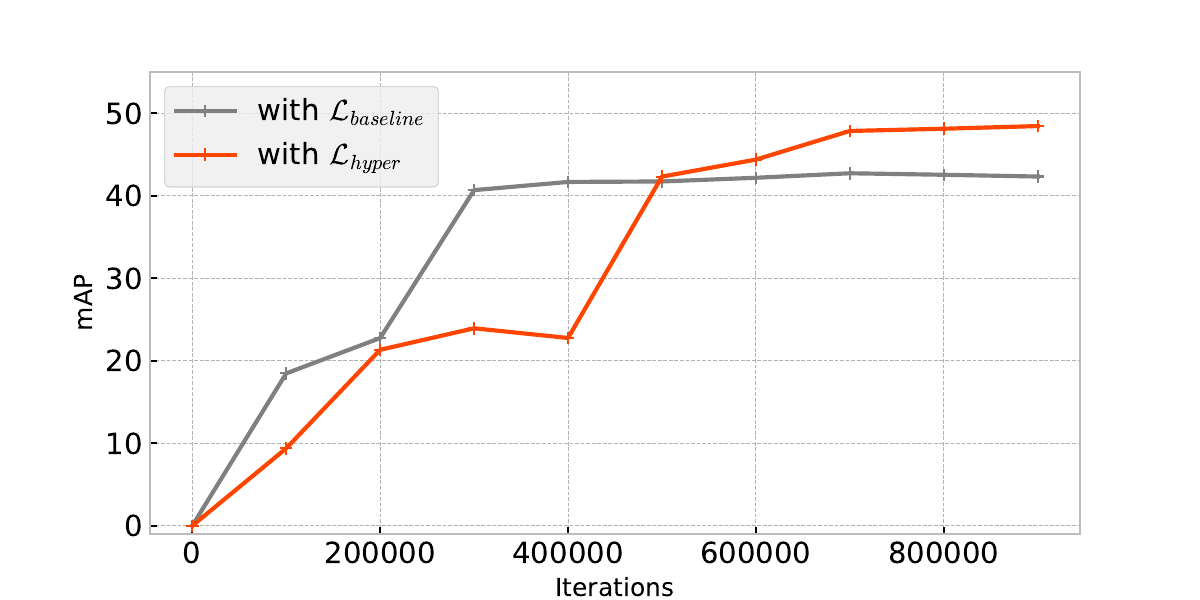}
    \vskip -0.8em
    \caption{Comparison between baseline objective Eq. \eqref{eq:baseline} and our objective Eq. \eqref{eq:ours} on COCO  during training. Metric: mAP. }
    \vskip -1em
    \label{fig:ap_over_iter}
\end{figure}

\begin{table}[!t]
    \centering
    \resizebox{0.44\textwidth}{!}{%
    \begin{tabular}{l @{\hskip 1mm}|l|c|c|c}
        \toprule
        & Method &  COCO & LVIS & ODinW \\ \midrule
        1& no synthetic captions &  42.3 & 13.3 & 26.3 \\ \midrule
        2& synthetic captions on $\mathcal{B}$ & 47.1  & 23.4 & 36.8  \\ 
        3& synthetic captions on $\mathcal{B}\cup\mathcal{G}$  & 47.6 & 23.7 & 38.6 \\ 
        4& synthetic captions on $\mathcal{B}\cup\mathcal{P}$ & 47.4 & 27.1 & 38.9 \\ 
        \rowcolor{LightCyan} 
        5& synthetic captions on $\mathcal{B}\cup\mathcal{G}\cup\mathcal{P}$ & \textbf{48.4} & \textbf{31.3} & \textbf{45.2} \\ 
        \bottomrule
    \end{tabular}
    }
    \vskip -0.8em
    \caption{{\bf Ablation study on region sampling for bootstrapping synthetic captions.} Metric: mAP.}
    \vskip -1.5em
\label{tab:ab_region_sampling}
\end{table}

\subsection{Ablation study}
\label{sec:ablation}

We now ablate our approach and its components: hyperbolic learning objective (\S \ref{sec:hyp_learn}), region sampling for bootstrapping synthetic captions (\S \ref{sec:syn_caps}), and cross-modal attention module for localization with free-form texts (\S \ref{sec:prelim}). 

\myparagraph{Ablation study on learning objectives.} Table \ref{tab:ab_loss} shows the results of different objectives: (1) standard open-world detection objective $\mathcal{L}_{Det}$ in Eq. \eqref{eq:basic}, (2) vanilla contrastive learning objective $\mathcal{L}_{baseline}$ in Eq. \eqref{eq:baseline}, (3) hyperbolic learning objective $\mathcal{L}_{hyper}$ in Eq. \eqref{eq:ours}. Our results show that our objective $\mathcal{L}_{hyper}$ outperforms other alternatives significantly, consistently across different datasets. The improvement of $\mathcal{L}_{baseline}$ over $\mathcal{L}_{Det}$ shows the strength of learning with synthetic captions to boost open-world generalization; while the improvement of $\mathcal{L}_{hyper}$ over $\mathcal{L}_{baseline}, \mathcal{L}_{Det}$ further shows the benefit of aligning visual and caption embeddings in hierarchy with hyperbolic geometry (Figure \ref{fig:loss}). 

Figure \ref{fig:ap_over_iter} compares the baseline objective $\mathcal{L}_{baseline}$ and our objective $\mathcal{L}_{hyper}$ during training. When using vanilla contrastive learning ($\mathcal{L}_{baseline}$), the performance saturates quickly. In contrast, our HyperLearner converges slower but consistently improves its performance, suggesting its stronger capability of learning from synthetic captions.

\begin{table}[!t]
    \centering
    \resizebox{0.43\textwidth}{!}{%
    \begin{tabular}{l @{\hskip 1mm}|l|c|c}
        \toprule
        &\multirow{2}{*}{Method}  &  RefCOCO &  \multirow{2}{*}{COCO} \\ 
        & &  val $ \ \ \  |$ testA \ $ |$ testB \\
        \midrule
        1& {\it w/o} cross-modal attention & 60.10 $|$ 70.96 $|$ 48.58 & \textbf{48.7} \\ 
       \rowcolor{LightCyan} 
        2& {\it w} cross-modal attention & \textbf{77.89 $|$ 76.92 $|$ 72.99} & 48.4 \\ 
        \bottomrule
    \end{tabular}
    }
    \vskip -0.8em
    \caption{{\bf Ablation study on cross-modal attention.}}
    \vskip -1em
\label{tab:vlfuse}
\end{table}

\begin{figure*}[!t]
    \centering
    \includegraphics[width=0.92\textwidth]{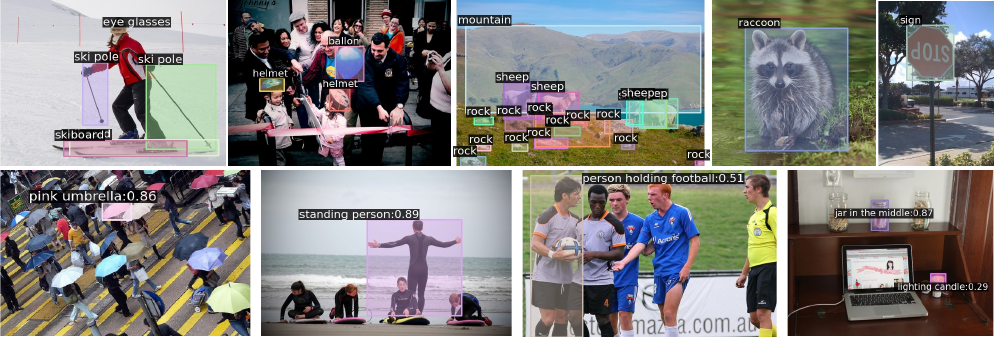}
    \vskip -0.8em
    \caption{
    {\bf Qualitative results on open-world detection.} Given any new class labels (row 1) or free-form texts (row 2) that specify objects with attribute, action, interaction, and spatial relationship, our model can detect and localize the objects in images.
    }
    \label{fig:qualitative}
    \vskip -1em
\end{figure*}

\myparagraph{Ablation study on region sampling.} Table \ref{tab:ab_region_sampling} shows the results of bootstrapping richer synthetic captions on different regions: 
basic groundtruth bboxes $\mathcal{B}$, proposal sampling bboxes $\mathcal{P}$, grid sampling bboxes $\mathcal{G}$ (\S \ref{sec:syn_caps}, Figure \ref{fig:model}). As shown in row 1 and 2, utilizing captions on $\mathcal{B}$ improves the performance substantially, with a $\approx$10\% increase on LVIS and ODinW. Moreover, the object-centric proposal sampling $\mathcal{P}$ (row 4) can target on novel objects, hence improving the detection on LVIS and ODinW, where performance are more dominated by rare novel objects. Finally, combining all sampled regions (row 5) yields the best overall performance, suggesting that using a diverse set of regions leads to richer semantics information and higher robustness. 

\myparagraph{Ablation study on cross-modal attention.} Table \ref{tab:vlfuse} shows the results of using cross-modal attention to fuse spatial visual information and language semantics  (Eq. \eqref{eq:attention}, \eqref{eq:spatial}, \eqref{eq:fuse} in \S \ref{sec:prelim}). Our results indicate that the cross-modal attention module improves our model performance significantly on RefCOCO benchmark and 
yields similar performance on COCO benchmark. These results prove the crucial effectiveness of leveraging spatial and language information for the task of localization with free-form texts, as evidenced by the substantial boosts of 17.79\%, 5.96\%, 24.31\% in Top-1 accuracy on the RefCOCO val, testA, testB respectively.

\begin{figure}[!t]
    \centering
    \includegraphics[width=\linewidth]{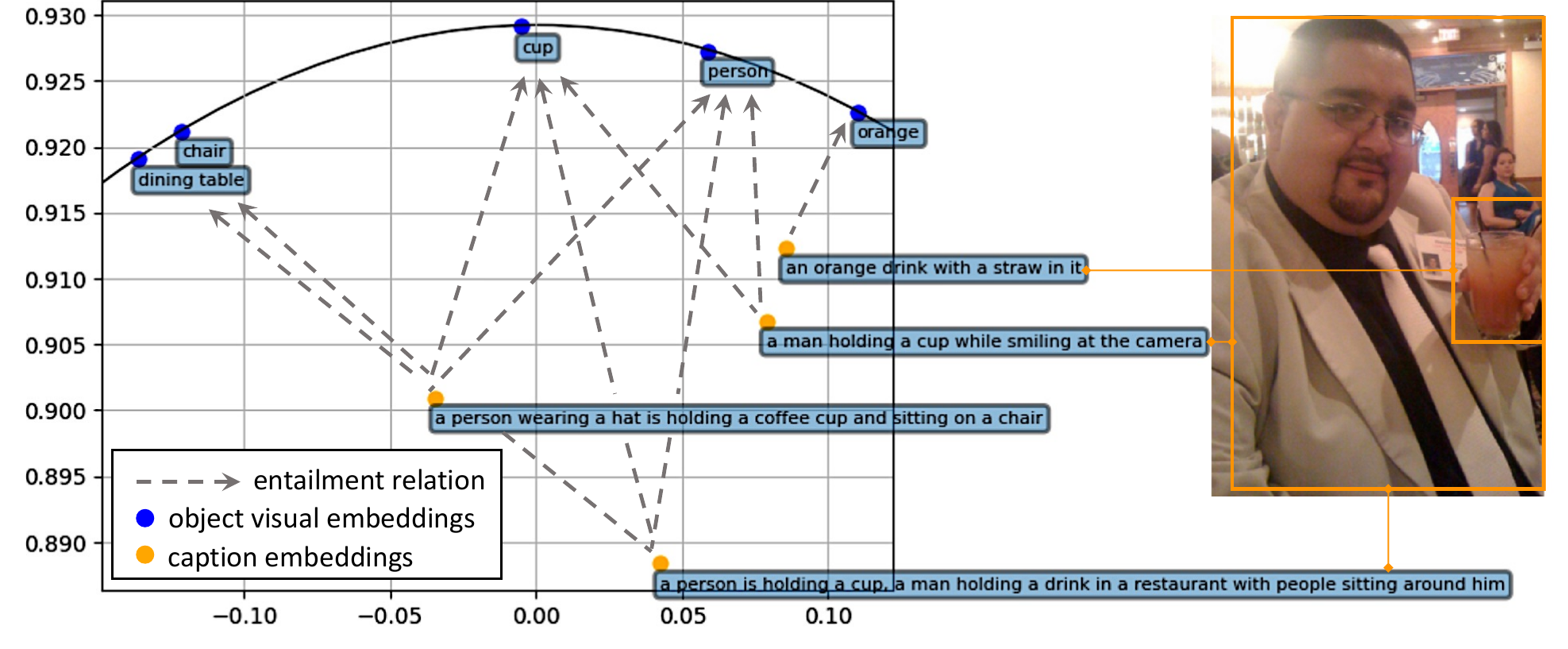}
    \vskip -0.8em
    \caption{{\bf Visualization of our object and caption embeddings.}}
    \vskip -1.5em
    \label{fig:vis_embeddings}
\end{figure}

\subsection{Qualitative analysis} 
\label{sec:qualitative}

\myparagraph{Qualitative results.}
Figure \ref{fig:qualitative} shows our qualitative results. 
In row 1, given new labels, our model can detect these new objects, even though these objects are small-scale (examples 2 and 3) and rare (example 4). In row 2, given any free-form texts that specify novel objects with fine-grained details, our model can localize these objects precisely, such as localizing the object of an attribute ({\em pink}),  performing an action ({\em standing}), interacting with other object ({\em holding football}), or in a certain spatial location ({\em in the middle}). These examples indicate the strong generalizability of our model on recognizing new concepts in open world.

\myparagraph{Visualization of embeddings.} Figure \ref{fig:vis_embeddings} shows our object visual embeddings, caption embeddings with UMAP \cite{mcinnes2018umap}, which projects the learned embeddings into a 2D space. We can find that the visualized embeddings present a tree-like entailment relation; this is in line with our motivation of imposing the hierarchy of {\em `caption entails object'} in our learning objective (Figure \ref{fig:loss}). Interestingly, we can also find the hierarchy of {\em `longer caption entails shorter caption'}, which suggests the intrinsic hierarchy inside the caption data.
\section{Conclusion}
\label{sec:conclusion}

We present a novel hyperbolic vision-language learning approach to learn from synthetic captions effectively. Our approach is well designed to leverage the open-world knowledge from pre-trained VLMs, and specially mitigate the hallucination issue in synthetic captions to boost the open-world generalizability in detection. Our comprehensive experiments on multiple benchmark datasets show the competitive performance of our approach as compared to the state-of-the-art. We analyze our model design rationales with insightful ablation study and qualitative analysis. 
Our work also paved a strong foundation on leveraging synthetic captions with hyperbolic learning for other vision tasks. 

\section*{\Large Supplementary Material}
\setcounter{figure}{0}
\setcounter{section}{0}
\setcounter{table}{0}
\renewcommand\thesection{\Alph{section}}
\renewcommand\thefigure{\Alph{figure}}
\renewcommand\thetable{\Alph{table}}

\section{Implementation details} 
\label{sec:implement}

In Table \ref{tab:pretraining}, we present the learning schedules when fine-tuning on different pre-training datasets. Note that when adding GoldG, or RefC in row 2, 3,  our model is initialized from the checkpoints in row 1, 2 respectively to continue model training. In Table \ref{tab:learning}, we present the learning schedules when fine-tuning on different downstream datasets. 
During both pre-training and fine-tuning, the learning rate is decayed by 0.1 at 1/3 and 2/3 of the training progress.

\begin{table}[!h]
  \centering
  \resizebox{0.9\linewidth}{!}{%
  \begin{tabular}{lllll}
    \toprule
    Model & Dataset(s) & Iterations & batch size & learning rate \\
    \midrule
    \multirow{3}{*}{HyperLearner} 
    & O365 & 900,000 & 64 & 0.0002 \\
    & O365,GoldG & +500,000 & 64 & 0.0002 \\
    & O365,GoldG,RefC & +500,000 & 64 & 0.0002 \\
    \bottomrule
  \end{tabular}
  }
  \vskip -1em
  \caption{Learning schedules on different pre-training datasets.}
\label{tab:pretraining}
\vskip -1em
\end{table}

\begin{table}[!h]
  \centering
  \resizebox{0.9\linewidth}{!}{%
  \begin{tabular}{lllll}
    \toprule
    Model & Dataset(s) & Iterations & batch size & learning rate \\
    \midrule
    \multirow{4}{*}{HyperLearner} 
    & COCO & 100,000 & 64 & 0.0001 \\
    & ODinW & 50,000 & 64 & 0.0001 \\
    & RefCOCO & 100,000 & 32 & 0.0001 \\
    \bottomrule
  \end{tabular}
  }
  \vskip -1em
  \caption{Learning schedules on different downstream datasets.}
\label{tab:learning}
\vskip -1em
\end{table}

\section{Computation cost of hyperbolic learning} 
\label{sec:computation}

To mitigate the high computational cost of hyperbolic learning, we make the following design choices. First, we choose to only lift the last-layer Euclidean features into hyperbolic space, instead of using a hyperbolic neural network. This approach circumvents computationally intensive operations, such as Frechet mean calculations, as noted in \cite{guo2022clipped}.

Second, we adopt the Lorentz model as  \cite{lou2020differentiating}, rather than the Poincaré ball model employed in \cite{desai2023hyperbolic}. The Poincaré ball model is operationally intensive due to its computationally heavy Gyrovector operations, which necessitate a series of intermediate calculations. In contemporary deep learning frameworks, these calculations substantially increase the overhead of computation graph, as described in \cite{ge2023hyperbolic}. In contrast, the Lorentz model offers a more computationally efficient alternative for hyperbolic geometry, which employs four-vector operations (Euclidean axes plus a time dimension), thus learning only one additional dimension. Notably, the time dimension can be readily determined based on the curvature of the hyperboloid, as detailed in \cite{desai2023hyperbolic}.

Third, thanks to the two design choices made above, the only additional computational costs are hyperbolic functions, including “sinh” and “cosh”. Fortunately, PyTorch efficiently optimizes these operations, particularly through techniques such as operation fusion. This technique combines multiple pointwise operations -- including arithmetic and trigonometric functions -- into a single kernel, thus further enhancing computational efficiency, as detailed in \cite{desai2023hyperbolic}.

\section{Discussion on evaluation benchmarks} 
\label{sec:discuss}

\myparagraph{Open-vocabulary detection evaluation benchmarks.}
It is worth noting that existing works in open-vocabulary detection (OVD) \cite{zareian2021open,gu2021open,wu2023aligning,zang2022open,zhong2022regionclip,wu2023cora,cho2023open,li2023blip} consider the evaluation setting where only {\em novel} ({\em unseen}) object classes are presented in the evaluation benchmarks. Specifically, the OVCOCO and OVLVIS benchmarks are built by splitting the class labels into two sets: seen and unseen. The model is trained on the seen classes and tested on the unseen classes.

\myparagraph{Open-world detection evaluation benchmarks.}
In contrast, our open-world detection task consider the evaluation setting where any object classes (both seen and unseen) are presented in the evaluation benchmarks, which aligns with detection in the open world. Moreover, these object classes could be described by keywords (class labels) or free-form texts. Our evaluation setting also poses a unique challenge of understanding novel concepts in free-form texts.

\section{Additional results} 
\label{sec:results}

\begin{table*}[!t]
  \centering
  \resizebox{\linewidth}{!}{%
  \begin{tabular}{l | c | c | ccccccccccccc c}
    \toprule
    Model & Evaluation Mode & Data & AerialDrone & Aquarium & Rabbits & EgoHands & Mushrooms & Packages & PascalVOC & pistols & pothole & Raccoon & Shellfish & Thermal & Vehicles & mAP \\
    \midrule
    \multirow{4}{*}{\makecell[l]{HyperLearner}} 
     & zero-shot & O365 &
      16.63 & 27.35 & 72.53 & 12.47 & 41.08 & 63.68 & 55.88 & 23.03 & 15.31 & 42.68 & 21.50 & 40.51 & 59.92 & 36.05 \\
     & fine-tune & O365 & 
     41.53 & 50.72 & 75.59 & 68.41 & 83.45 & 72.34 & 84.39 & 99.77 & 52.08 & 61.00 & 43.09 & 75.35 & 59.42 & 67.31 \\
     & zero-shot & O365, GoldG & 
     27.25 & 35.34 & 74.38 & 18.12 & 58.50 & 68.95 & 64.12 & 27.03 & 17.06 & 56.01 & 29.36 & 60.72 & 60.24 & 44.74 \\
     & fine-tune & O365, GoldG & 
     42.48 & 52.10 & 76.34 & 74.66 & 90.43 & 76.53 & 87.94 & 99.77 & 53.00 & 63.58 & 39.52 & 78.52 & 61.25 & 69.57 \\
    \bottomrule
  \end{tabular}
  }
  \vskip -0.5em
  \caption{{\bf Detailed results of zero-shot and fine-tune transfer on 13 individual datasets on ODinW.} 
  Metric: mAP over 13 datasets.
  }
  \label{tab:odinw13}
\vskip -1em
\end{table*}

\begin{figure*}[!t]
    \centering
    \includegraphics[width=0.99\textwidth]{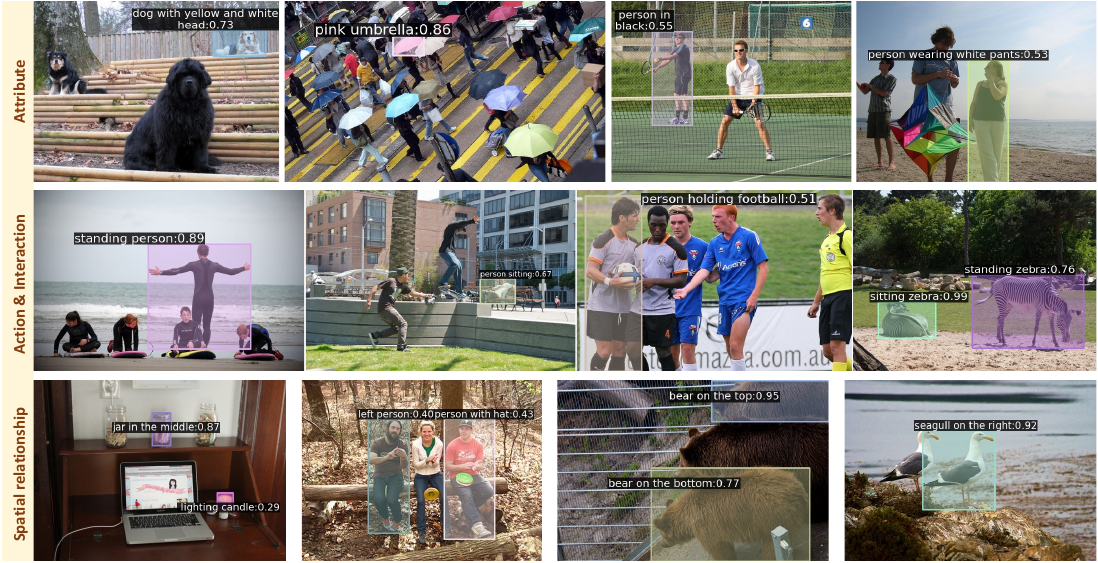}
    \vskip -0.8em
    \caption{
    {\bf More qualitative results on open-world detection using free-form texts,} where the free-form texts specify objects with attribute, action/interaction, and spatial relationships. 
    }
    \label{fig:more_qualitative}
    \vskip -1em
\end{figure*}

\myparagraph{Additional quantitative results on ODinW.} Table \ref{tab:odinw13} shows the additional results of 13 individual datasets on ODinW benchmark of our proposed approach, when using different evaluation modes (zero-shot or fine-tune/full-shot) and different pre-training data. Results show the benefit of using more data for pre-training, and further fine-tuning on different target datasets to improve generalization.  

\myparagraph{Additional qualitative results.} Figure \ref{fig:more_qualitative} presents more qualitative examples using our model to localize objects described using free-form texts to specify certain object attribute, action/interaction, and spatial relationships. Examples illustrate the strong generalizability of our model in understanding and recognizing novel concepts in open world.

\section{Discussion on quality of synthetic captions}
\label{sec: discuss_caption}
Generally, our VLM, BLIP2 6.7b, generates synthetic captions that accurately describe the objects in the image, as indicated by \cite{li2023blip}. Furthermore, our experimental result reveals that the synthetic captions generated by VLM can significantly improve the performance of our model across various open-world object detection benchmarks. These observations qualitatively demonstrates the quality of synthetic captions for training open-world object detection models. In this section, we carry out a quantitative study for the quality of synthetic captions generated by VLMs, by determining the ratio of noisy or hallucinated objects in generated captions.

To quantify the noise level in our synthetic captions (i.e., \% incorrect objects described), we adopt the Caption Hallucination Assessment with Image Relevance (CHAIR) metric from \cite{rohrbach2018object,li2023evaluating} as our noise-level measurement for synthetic captions. It can be formulated as the percentage of incorrectly described objects,

\begin{equation}
    \text{noise \%} = \frac{1}{m}\sum_{i=1}^m \frac{N_{\text{incorrect objects}}}{N_{\text{total objects}}} \times 100\%
\label{eq:cap_noise}
\end{equation}
,where $N_{\text{incorrect objects}}$ denotes the number of hallucinated objects in the caption, $N_{\text{total objects}}$ denotes the number of total objects in the caption, and $m$ is the total number of synthetic captions collected from the dataset. To obtain $N_{\text{incorrect objects}}$ and $N_{\text{total objects}}$ in (Eq. \eqref{eq:cap_noise}), we detect the synonyms of object classes within the synthetic captions using WordNet synsets~\cite{miller1995wordnet}. If an object or its synonym is mentioned in the caption, but the detection ground truth does not include this object, it is classified as an incorrect object. Our result indicates a noise level of 16.3\% on COCO 2017 test set, using BLIP2 6.7b. Despite the presence of noise and hallucinated contents in the synthetic captions, we have proposed a hyperbolic learning strategy to mitigate the issue of training with noisy synthetic captions. The hyperbolic learning loss learns visual and caption embeddings hierarchically, imposing an entailment relationship between captions and region embeddings, rather than directly aligning them. Moreover, we suggest that employing an enhanced pre-trained VLM for captioning could further mitigate the noise level.

\section{Future work} 
\label{sec:future_work}

Building upon our current work in open-world object detection, there are several promising avenues for future exploration. First, it is essential to develop a new benchmark that evaluates open-world object detection using both keywords and free-form text, overcoming the limitations of current benchmarks that merely focus on either keywords (e.g., COCO, LVIS, ODinW)  or free-form texts (e.g., RefCOCO). Second, training image captioning models that generate object-centric synthetic captions presents an intriguing research direction to provide better open-world knowledge in object detection. Third, integrating various data modalities or information sources to augment the model's detection capabilities is also an interesting direction, such as providing a sketch of a target object to localize in the open world. 

\clearpage
\newpage
{
    \small
    \bibliographystyle{ieee_fullname}
    \bibliography{reference}
}

\end{document}